\pdfoutput=1
\documentclass[10pt,twocolumn,letterpaper]{article}

\usepackage{cvpr}
\usepackage{times}
\usepackage{epsfig}
\usepackage{graphicx}
\usepackage{amsmath}
\usepackage{amssymb}

\usepackage[breaklinks=true,bookmarks=false]{hyperref}

\cvprfinalcopy 

\def\cvprPaperID{3435} 

\begin{document}
\setlength{\textfloatsep}{10pt plus 1.0pt minus 2.0pt}

\title{CleanNet: Transfer Learning for Scalable Image Classifier Training \newline with Label Noise\vspace{-0.7em}}

\author{Kuang-Huei Lee\textsuperscript{1} \qquad Xiaodong He\textsuperscript{2}\thanks{Work performed while working at Microsoft.} \qquad Lei Zhang\textsuperscript{1} \qquad Linjun Yang\textsuperscript{3}\footnotemark[1]\\
\textsuperscript{1}Microsoft AI and Research \qquad \textsuperscript{2}JD AI Research \qquad \textsuperscript{3}Facebook\\
{\tt\small \{kualee,leizhang\}@microsoft.com \qquad xiaodong.he@jd.com \qquad linjuny@fb.com}
}

\maketitle

\begin{abstract}
\vspace{-2.5mm}
In this paper, we study the problem of learning image classification models with label noise. Existing approaches depending on human supervision are generally not scalable as manually identifying correct or incorrect labels is time-consuming, whereas approaches not relying on human supervision are scalable but less effective. To reduce the amount of human supervision for label noise cleaning, we introduce CleanNet, a joint neural embedding network, which only requires a fraction of the classes being manually verified to provide the knowledge of label noise that can be transferred to other classes. We further integrate CleanNet and conventional convolutional neural network classifier into one framework for image classification learning. We demonstrate the effectiveness of the proposed algorithm on both of the label noise detection task and the image classification on noisy data task on several large-scale datasets. Experimental results show that CleanNet can reduce label noise detection error rate on held-out classes where no human supervision available by 41.5\% compared to current weakly supervised methods. It also achieves 47\% of the performance gain of verifying all images with only 3.2\% images verified on an image classification task. Source code and dataset will be available at kuanghuei.github.io/CleanNetProject.
\end{abstract}
\vspace{-2.5mm}

\section{Introduction}
\label{Introduction}
One of the key factors that drive recent advances in large-scale image recognition is massive collections of labeled images like ImageNet \cite{deng2009imagenet} and COCO \cite{lin2014microsoft}. However, it is normally expensive and time-consuming to collect large-scale manually labeled datasets. In practice, for fast development of new image recognition tasks, a widely used surrogate is to automatically collect noisy labeled data from Internet \cite{fergus2010learning,krause2016unreasonable,schroff2011harvesting}. Yet many studies have shown that label noise can affect accuracy of the induced classifiers significantly \cite{frenay2014classification,nettleton2010study,rolnick2017deep,sukhbaatar2014training}, making it desirable to develop algorithms for learning in presence of label noise.

Learning with label noise can be categorized by type of supervision: methods that rely on human supervision and methods that do not. For instance, some of the large-scale training data were constructed using classifiers trained on manually verified seed images to remove label noise (e.g. LSUN \cite{yu2015lsun} and Places \cite{zhou2017places}). Some studies for learning convolutional neural networks (CNNs) with noise also rely on manual labeling to estimate label confusion \cite{patrini2017making,xiao2015learning}. The methods using human supervision exhibit a disadvantage in scalability as they require labeling effort for every class. For classification tasks with millions of classes \cite{dean2013fast,frome2013devise}, it is infeasible to have even one manual annotation per class. In contrast, methods without human supervision (e.g. model predictions-based filtering \cite{frenay2014classification} and unsupervised outliers removal \cite{liu2014unsupervised,scholkopf2001estimating,xia2015learning}) are scalable but often less effective and more heuristic. Going with any of the existing approaches, either all the classes or none need to be manually verified. It is difficult to have both scalability and effectiveness.

In this work, we strive to reconcile this gap. We observe that one of the key ideas for learning from noisy data is finding ``class prototypes'' to effectively represent classes. Methods learn from manually verified seed images like \cite{yu2015lsun} and methods assume majority correctness like \cite{azadi2016auxiliary} belong to this category. Inspired by this observation, we develop an attention mechanism that learns how to select representative seed images in a reference image set collected for each class with supervised information, and transfer the learned knowledge to other classes without explicit human supervision through transfer learning. This effectively addresses the scalability problem of the methods that rely on human supervision. 

\begin{figure}[t]
\begin{center}
\includegraphics[width=1.0\linewidth]{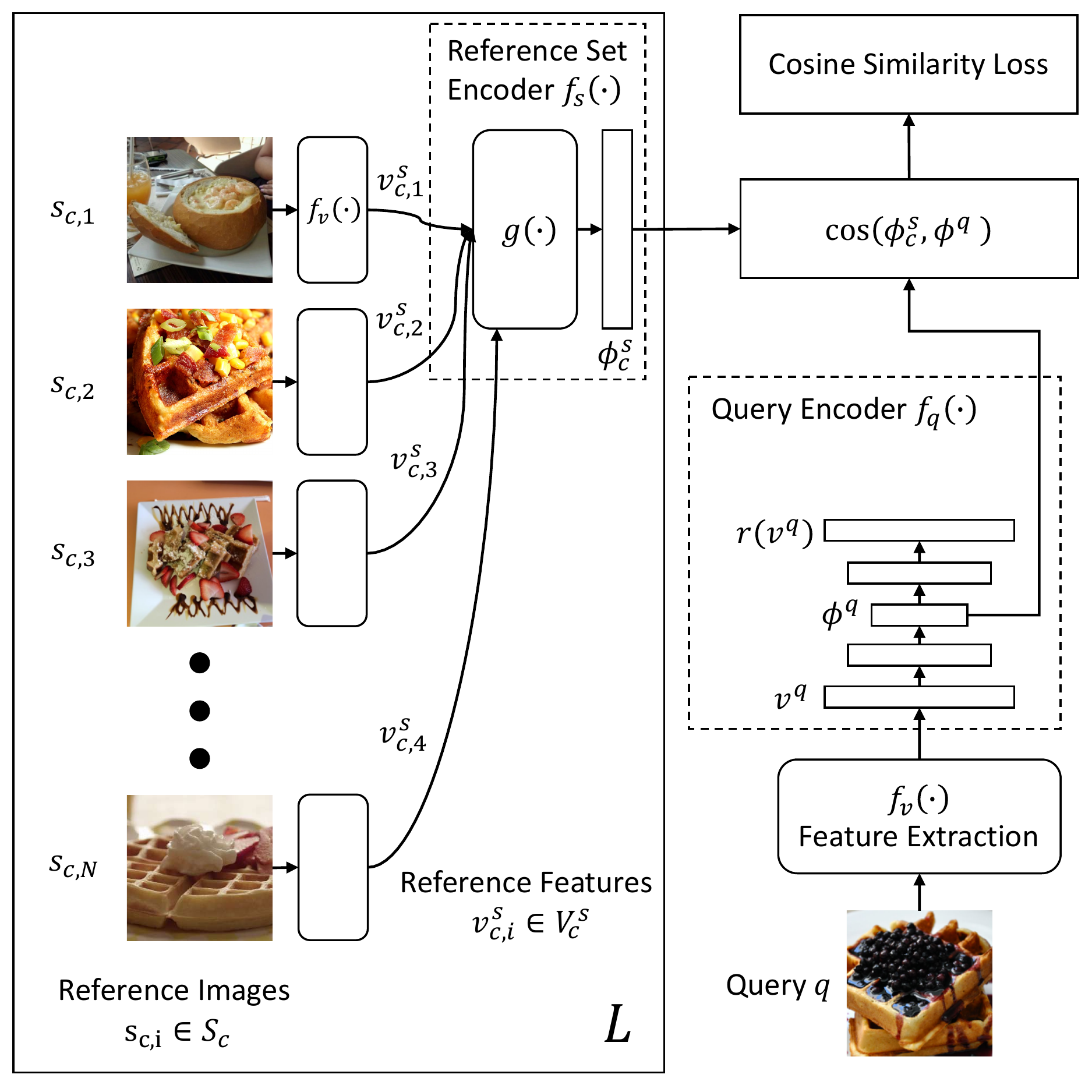}
\end{center}
\vspace{-3mm}
\caption{CleanNet architecture for learning a class embedding vector $\phi_c^s$ and a query embedding vector $\phi_q$ with a similarity matching constraint. There exists one class embedding for each of the $L$ classes. Details of component $g(\cdot)$ are depicted in Fig. \ref{fig:set_encoder}.}
\label{fig:cleansenet}
\end{figure}

Thus, we introduce ``label cleaning network'' (CleanNet), a novel neural architecture designed for this setting. First, we develop a reference set encoder with the attention mechanism to encode a set of reference images of a class to an embedding vector that represents that class. Second, in parallel to reference set embedding, we also build a query embedding vector for each individual image and impose a matching constraint in training to require a query embedding to be similar to its class embedding if the query is relevant to its class. In other words, the model can tell whether an image is mislabeled by comparing its query embedding with its class embedding. Since class embeddings generated from different reference sets represents different classes where we wish the model to adapt to, CleanNet can generalize to classes without explicit human supervision. Fig. \ref{fig:cleansenet} illustrates the end-to-end differentiable model. 

As the first step of this work, we demonstrate that CleanNet is an effective tool for label noise detection. Simple thresholding based on the similarity between the reference set and the query image lead to good results compared with existing methods. Label noise detection not only is useful for training image classifiers with noisy data, but also has important values in applications like image search result filtering and linking images to knowledge graph entities.

CleanNet predicts the relevance of an image to its noisy class label. Therefore, we propose to use CleanNet to assign weights to image samples according to the image-to-label relevance to guide training of the image classifier. On the other hand, as a better classifier provides more discriminative convolutional image features for learning CleanNet, we refresh the CleanNet using the newly trained classifier. We introduce a unified learning scheme to train the CleanNet and image classifier jointly.

To summarize, our contributions include a novel neural architecture CleanNet that is designed to make label noise detection and learning from noisy data with human supervision scalable through transfer learning. We also propose a unified scheme for training CleanNet and the image classifier with noisy data. We carried out comprehensive experimentation to evaluate our method for label noise detection and image classification on three large datasets with real-world label noise: Clothing1M \cite{xiao2015learning}, WebVision \cite{li2017webvision}, and Food-101N. Food-101N contains 310K images we collected from Internet with the Food-101 taxonomy \cite{bossard2014food}, and we added ``verification label'' that verifies whether a noisy class label is correct for an image\footnote{Food-101N will be available at kuanghuei.github.io/CleanNetProject.}. Experimental results show that CleanNet can reduce label noise detection error rate on held-out classes where no human supervision available by 41.5\% compared to current weakly supervised methods. It also achieves 47\% of the performance gain of verifying all images with only 3.2\% images verified on an image classification task.

\section{Related Work}
\noindent
\textbf{Label noise reduction.} Our method belongs to the category of approaches that address label noise by demoting or removing mislabeled instances in training data. One of the popular approaches is unsupervised outlier removal (e.g. One-Class SVM \cite{scholkopf2001estimating}, UOCL \cite{liu2014unsupervised}, and DRAE \cite{xia2015learning}). Using this approach for label noise detection relies on an assumption that outliers are mislabeled. However, outliers are often not well defined, and therefore removing them presents a challenge \cite{frenay2014classification}. Another approach that also needs no human supervision is weakly supervised label noise reduction \cite{frenay2014classification}. For example, Thongkam \etal \cite{thongkam2008support} proposed a classification filtering method that learns an SVM from noisy data and removes instances misclassified by the SVM. Weakly supervised methods are often heuristic, and we are not aware of any large dataset actually built with these methods. On the other hand, label noise reduction using human supervision has been widely studied for dataset constructions. For instance, Yu \etal \cite{yu2015lsun} proposed manually labeling seed images and then training multilayer perceptrons (MLPs) to remove mislabeled images. Similarly, the Places dataset \cite{zhou2017places} was constructed using an AlexNet \cite{krizhevsky2012imagenet} trained on manually verified seed images. However, methods using human supervision exhibit a disadvantage in scalability as they require human supervision for every class to be cleansed. 

\noindent
\textbf{Direct neural network learning with label noise.} Some methods were developed for directly learning neural network with label noise \cite{azadi2016auxiliary,chen2015webly,li2017learning,patrini2017making,rolnick2017deep,sukhbaatar2014training,veit2017learning,xiao2015learning,zhuang2017attend}. Azadi \etal \cite{azadi2016auxiliary} developed a regularization method to actively select image features for training, but it depends on features pre-trained for other tasks and hence is less effective. Zhuang \etal \cite{zhuang2017attend} proposed attention in random sample groups but did not compare with standard CNN classifiers, and thus is less practical. Methods proposed by Xiao \etal \cite{xiao2015learning} and Patrini \etal \cite{patrini2017making} rely on manual labeling to estimate label confusion for real-world label noise. However, such labeling is required for all classes and much more expensive than simply verifying whether the noisy class labels are correct. Veit \etal \cite{veit2017learning} proposed an architecture that learns from human verification to clean noisy labels, but their approach does not generalize to classes that are not manually verified as opposed to our method. Chen \etal \cite{chen2015webly}, which relies on specific data sources, and Li \etal \cite{li2017learning}, which uses knowledge graph, could be difficult to generalize and thus are beyond the scope of this paper.


\noindent
\textbf{Transfer learning with neural network.} There is a large body on literature of learning neural joint embeddings for transfer learning \cite{frome2013devise,salvador2017learning,socher2013zero,tsai2017learning,vinyals2016matching}. Tsai \etal \cite{tsai2017learning} trained visual-semantic embeddings with supervised and unsupervised objectives using labeled and unlabeled data to improve robustness of embeddings for transfer learning. Recently Liu \etal \cite{liu2016coupled} and Tzeng \etal \cite{tzeng2017adversarial} exploited adversarial objectives for domain adaptation. Inspired by \cite{tsai2017learning}, we also incorporate unsupervised objectives in this work.

\section{Scalable Learning with Label Noise}
We focus on learning an image classifier from a set of images with label noise using transfer learning. Specifically, assume we have a dataset of $n$ images, i.e., $X = \{(x_1,y_1),...,(x_n,y_n)\}$, where $x_i$ is the $i$-th image and $y_i \in \{1,...,L\}$ is its class label, where $L$ is the total number of classes. Note that the class labels are noisy, means some of the images' labels are incorrect.

In this section, we present the CleanNet, a joint neural embedding network, which only requires a fraction of the classes being manually verified to provide the knowledge of label noise that can be transferred to other classes. We then integrate CleanNet and conventional convolutional neural network (CNN) into one system for image classifier training with label noise.  Specifically, we introduce the designs and properties of CleanNet in Section \ref{sec:denoisenet}. In Section \ref{sec:combine} we integrate CleanNet and the CNN into one framework for image classifier learning from noisy data.

\subsection{CleanNet}
\label{sec:denoisenet}
The overall architecture of CleanNet is shown in Fig. \ref{fig:cleansenet}. It consists of two parts: a reference set encoder and a query encoder. The reference set encoder $f_s(\cdot)$ learns to focus on representative features in a noisy reference image set, which is collected for a specific class, and outputs a class-level embedding vector. Since using all the images in the reference set is computationally expensive, we first create a representative subset, and extract one visual feature vector from each image in that subset to form a representative feature vector set, i.e., let $V_c^s$ denotes the representative reference feature vector set for class $c$ (reference feature set). 

We explored two pragmatic approaches to select $V_c^s$. The first one is random sampling a subset from all images in class $c$ and extract features using a pre-trained CNN $f_v(\cdot)$ as shown in Fig. \ref{fig:cleansenet}. The second approach is running K-means on the extracted features of all images in class $c$ to find K cluster centroids and use them as $V_c^s$. The K-means step is ignored in the figures. Since the K-means approach shows slightly better result on a held-out set, we choose it for experiments hereafter. We select 50 feature vectors to form $V_c^s$.

In parallel to reference set encoder, we also develop a query encoder $f_q(\cdot)$. Let $q$ denote a query image labeled as class $c$. The query encoder $f_q(\cdot)$ maps the query image feature $v^q = f_v(q)$ to a query embedding $\phi^q = f_q(v^q)$. We impose a matching constraint such that the query embedding $\phi^q$ is similar to its class embedding $\phi_c^s = f_s(V_c^s)$ if the query $q$ is relevant to its class label $c$. In other words, we decide whether a query is mislabeled by comparing its query embedding vector with its class embedding vector. Since the class labels are noisy, we can further mark up a query image and its class label by a manual ``verification label''. The verification label for each image is defined as 
\begin{equation}
\small
l = 
\begin{cases}
    1& \text{if the image is relevant to its noisy class label}\\
    0& \text{if the image is mislabeled}\\
    -1 & \text{if verification label not available}
\end{cases}
\end{equation}
Note that, to reduce human labeling effort, most of the verification labels are -1, means no human verification available.

The model learns the matching constraint from the supervision given by the verification labels, such that a query embedding is similar to its class embedding if the query image $q$ truly belongs to its class label, and transfer to different classes where no human verification available. In the following, we present how we build the reference set encoder, query encoder, and objectives for learning the matching constraint.

\begin{figure}[t]
\begin{center}
\includegraphics[width=1.0\linewidth]{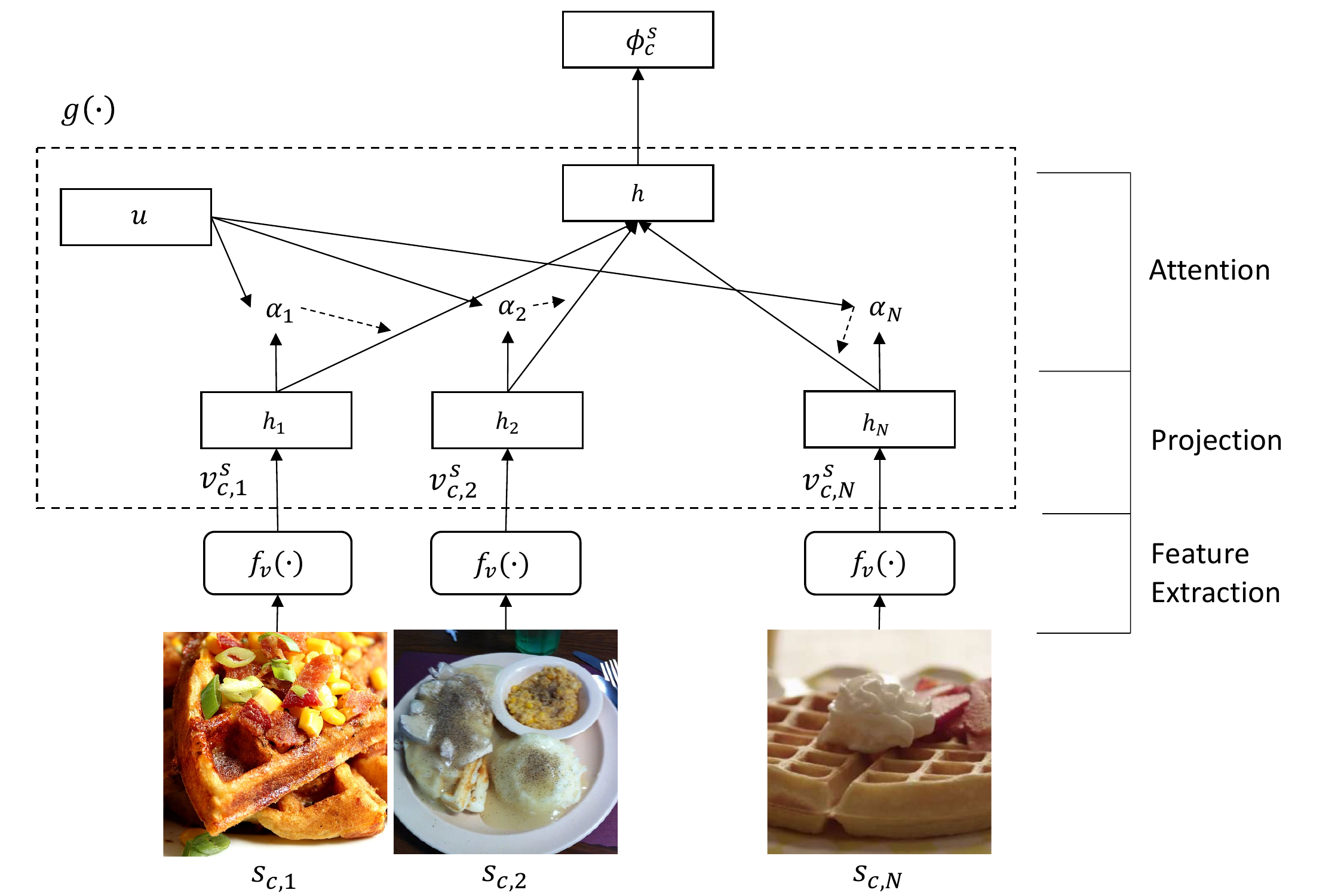}
\end{center}
\vspace{-3mm}
\caption{Reference set encoder $f_s(\cdot)$}
\label{fig:set_encoder}
\end{figure}

\begin{figure}[t]
\begin{center}
\includegraphics[width=1.0\linewidth]{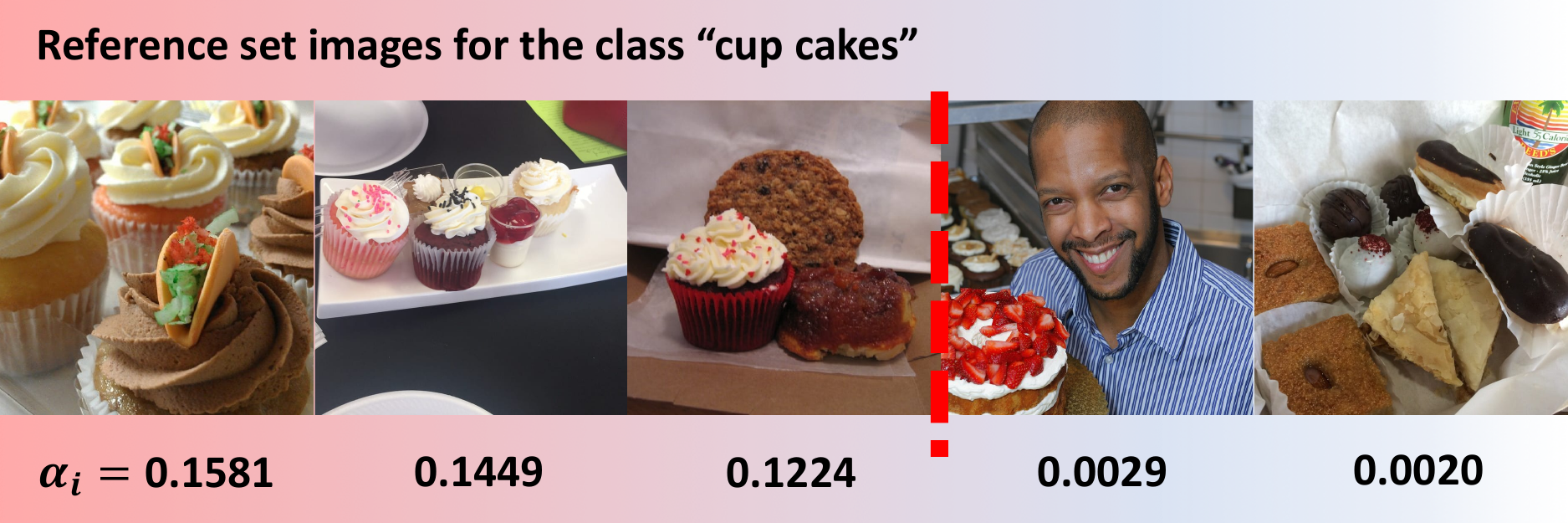}
\end{center}
\vspace{-3mm}
\caption{Examples that received the most and the least attention in a reference set for "cup cakes". $\alpha_i$ is defined in Eq. \eqref{eq:alpha}. }
\label{fig:att}
\end{figure}

\noindent
\textbf{Reference set encoder.} The architecture of the reference set encoder is depicted in Fig. \ref{fig:set_encoder}. It maps a reference feature set $V_c^s$ for class $c$ to a class embedding vector $\phi_c^s$. First, a two-layer MLP projects each image feature to a hidden representation $h_i$. Next, we learn an attention mechanism to encode representative features to a fixed-length hidden representation as class prototype:
\begin{equation}
u_i = tanh(Wh_i + b)
\end{equation}
\begin{equation}
\alpha_i = \dfrac{exp(u_i^Tu)}{\sum_iexp(u_i^Tu)}
\label{eq:alpha}
\end{equation}
\begin{equation}
h = \sum_i\alpha_ih_i
\label{eq:attention_avg}
\end{equation}
As shown in Eq. \eqref{eq:attention_avg}, the importance of each $h_i$ is measured by the similarity between $u_i$ and a context vector $u$. Similar to \cite{yang2016hierarchical}, the context vector $u$ is learned during training. Driven by the matching constraint, this attention mechanism learns how to pay attention on the most representative features for classes. This model learns from supervised information, i.e., the manual verification label, and adapts to other classes without explicit supervision. An example of this attention mechanism is shown in Fig. \ref{fig:att}. Finally, a one-layer MLP maps the hidden representation to the class embedding $\phi_c^s$.

\noindent
\textbf{Query encoder.} As illustrated in Fig. \ref{fig:cleansenet}, we adopt a 5-layer autoencoder \cite{hinton2006reducing} as the query encoder and incorporate autoencoder reconstruction error into learning objectives. Taking this strategy, as proposed in \cite{tsai2017learning}, forces the query embedding to preserve semantic information of all the classes including those classes without verification labels, because images without verification label can now be used in training with this unsupervised objective. It has been proven effective for improving domain adaptation performance.

Given a query image feature vector $v^q$, the autoencoder maps $v^q$ to a hidden representation $\phi^q$ and seek to reconstruct $v^q$ from $\phi^q$. The reconstruction error is defined as
\begin{equation}
L_{r}(v^q) = {||v^q-r(v^q)||}^2
\label{autoencoder}
\end{equation}
where $r(v^q)$ is the reconstructed representation.

\noindent
\textbf{Learning objectives based on matching constraint.} With the supervision from human verification labels, the similarity between class embedding $\phi_c^s$ and query embedding $\phi^q$ is maximized if a query is relevant to its class label ($l=1$); otherwise the similarity is minimized ($l=0$). We adopt the cosine similarity loss with margin to impose this constraint:
\begin{equation}
\small
L_{cos}(\phi^q, \phi_c^s, l) = 
\begin{cases}
    1-\cos(\phi^q, \phi_c^s)& \text{if } l=1\\
    \omega(max(0,\cos(\phi^q, \phi_c^s)-\rho))& \text{if } l=0 \\
    0 & \text{if } l=-1
\end{cases}
\label{eq:cos_sim}
\end{equation}
where $\cos(\cdot)$ is the normalized cosine similarity, $\omega$ is negative sample weight for balancing positive and negative samples, and $\rho$ is the margin set to 0.1 in this work. The case $l=-1$ is ignored in the loss function since this supervised objective only utilizes query images with verification label.

On the other hand, images without verification label can also be utilized to learn the matching constraint. Similar to \cite{tsai2017learning}, we introduce an unsupervised self-reinforcing strategy that applies pseudo-verification to images without verification label. To be specific, a query is treated as relevant if $\cos(\phi^q, \phi_c^s)$ is larger than the margin $\rho$: 
\begin{equation}
\small
L_{cos}^{unsup}(\phi^q, \phi_c^s) = 
\begin{cases}
    1-\cos(\phi^q, \phi_c^s)& \text{if } l_{sudo}=1\\
    0& \text{if } l_{sudo}=0
\end{cases}
\label{eq:selfreinforcing}
\end{equation}
\begin{equation}
\small
l_{sudo} = 
\begin{cases}
    1& \text{if } \cos(\phi^q, \phi_c^s) \geq \rho\\
    0& otherwise
\end{cases}
\label{eq:sudoverification}
\end{equation}
where $\rho$ is the same margin as in Eq. \eqref{eq:cos_sim}. From Eq. \eqref{eq:selfreinforcing} and Eq. \eqref{eq:sudoverification}, we can see that for queries that are initially treated as relevant, the model learns to further push up the similarity between queries and reference sets; for queries that are initially treated as irrelevant, they are ignored.

\noindent
\textbf{Total loss.} To summarize the training objectives, our model is learned by minimizing a total loss combining both supervised and unsupervised objectives:
\begin{equation}
L_{total} = L_{cos} + \beta L_{r} + t \gamma L_{cos}^{unsup}
\end{equation}
\begin{equation}
t = 
\begin{cases}
    1& \text{if } l = -1 \\
    0& \text{if } l \in \{0,1\}
\end{cases}
\end{equation}
where $\beta$ and $\gamma$ are selected through hyper-parameter search, and $t$ indicates whether a query image has verification label. $\beta$ and $\gamma$ are set to 0.1 in this work. During training, we randomly sample images without verification label as queries for a fraction of a mini-batch (usually $1/2$).

Note that the parameters of the attentional reference set encoder and the query encoder are tied across all classes so the information learned from classes that have human verification labels can be transferred to other classes that have no human verification label.

\subsection{CleanNet for Label Noise Detection.}
\label{sec:label-noise-detection}
From a relevance perspective, CleanNet can be used to rank all the images with label noise for a class by cosine similarity $\cos(\phi^q, \phi_c^s)$. We can simply perform thresholding for label noise detection:
\begin{equation}
\hat{l} = 
\begin{cases}
    1& \text{if } \cos(\phi^q, \phi_c^s) \geq \delta \\
    0& otherwise
\end{cases}
\label{eq:threshold}
\end{equation}
where $\delta$ is a threshold selected through cross-validation. We observe that the threshold is not very sensitive to different classes in most cases, and therefore we usually select an uniform threshold for all classes so that verification labels are not required for all classes for cross-validation.

\subsection{CleanNet for Learning Classifiers}
\label{sec:combine}

CleanNet predicts the relevance of an image to its noisy class label by comparing the query embedding of the image to its class embedding that represents the class. That is, the distance between two embeddings can be used to decide how much attention we should pay to a data sample in training the image classifier. Specifically, we assign attention weights on data samples based on the cosine similarity:
\begin{equation}
\small
w_{soft}(x, y=c, V_c^s) = max(0,\cos(f_q(f_v(x)), f_s(V_c^s)))
\label{eq:softweight}
\end{equation}
where $V_c^s$ is the reference image feature set that represents the prototype of class $y=c$. Eq. \eqref{eq:softweight} defines a soft weighting on an image $x$ with noisy class label $y=c$. Similarly, we also define a hard weighting as
\begin{equation}
w_{hard}(x, y=c, V_c^s) = 
\small
\begin{cases}
    1& \text{if } \cos(f_q(f_v(x)), f_s(V_c^s)) \geq \delta \\
    0& otherwise
\end{cases}
\label{eq:hardweight}
\end{equation}
where $\delta$ is a threshold as in Eq. \eqref{eq:threshold}. In essence, hard weighting is equivalent to explicit label noise removal. With $w_{soft}$ or $w_{hard}$, we define the weighted classification learning objective as 
\begin{equation}
\small
L_{weighted}(x, y=c, V_c^s) = w_{soft|hard}(x, y, V_c^s)H(x, y=c)
\label{eq:weighted_loss}
\end{equation}
where $H(x, y=c)$ is negative log likelihood:
\begin{equation}
H(x, y=c) = -\sum_{c=0}^Lp(y=c|x)log\hat{p}(y=c|x)
\end{equation}

\begin{figure}[t]
\begin{center}
\includegraphics[width=1.0\linewidth]{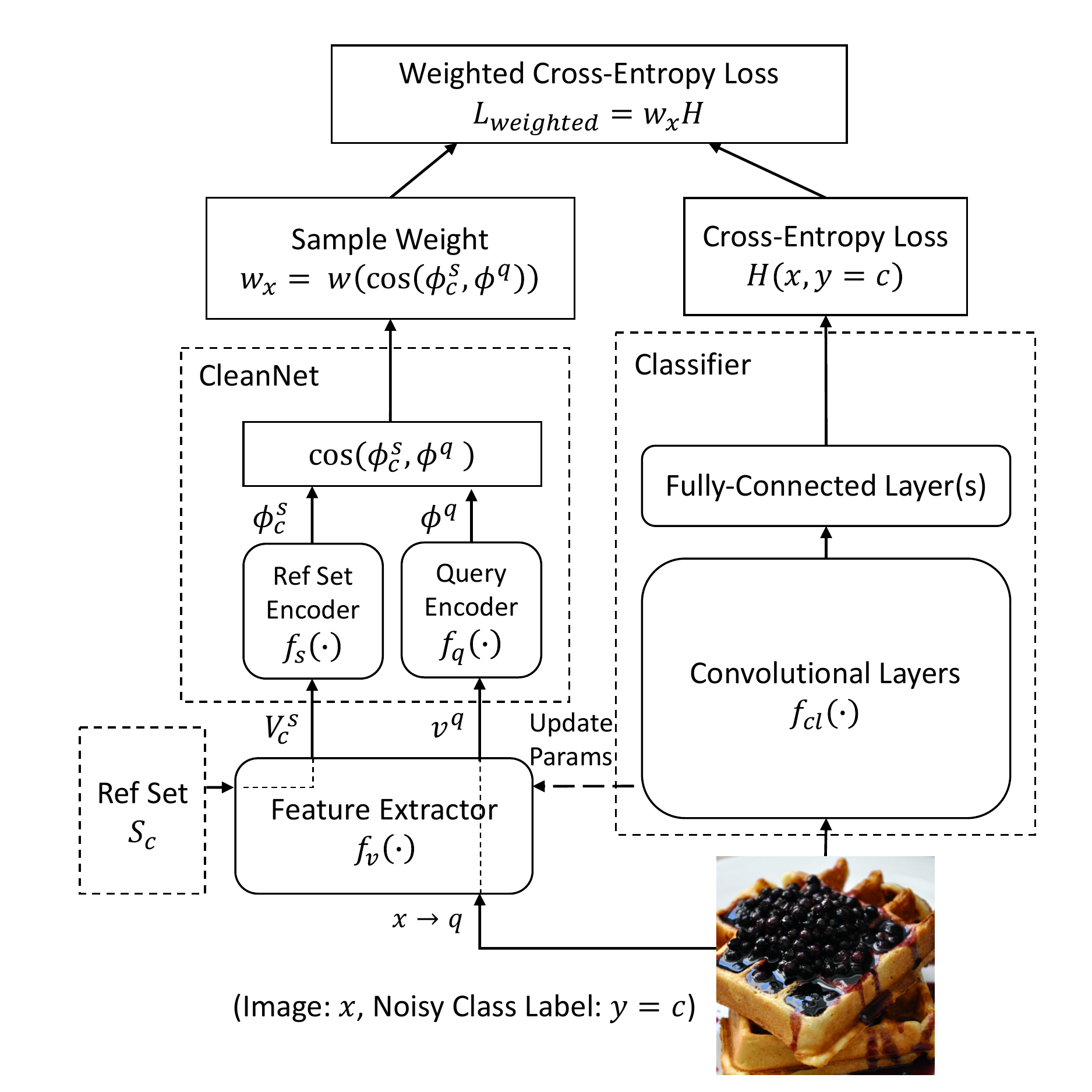}
\end{center}
\vspace{-3mm}
\caption{Illustration of integrating CleanNet for training the CNN-based image classifier with label noise.}
\label{fig:unified}
\end{figure}

\noindent
\textbf{Integrating CleanNet and the image classifier.}
Learning the image classifier relies on CleanNet to assign proper attention weights to data samples. On the other hand, better classifier provides more discriminative features which are critical for CleanNet learning. Therefore, we integrate CleanNet and the CNN-based image classifier into one framework for end-to-end learning of image classifiers with label noise. The overall architecture of this framework is illustrated in Fig. \ref{fig:unified}. The structure of a CNN-based image classifier is split into fully-connected layer(s) and convolutional layers $f_{cl}$ that can be used for feature extraction.

\noindent
\textbf{Alternating training.}
We adopt an alternative training scheme to learn the proposed classification system. At step 1, we first train a classifier from noisy data with all sample weights set to 1. At step 2, parameters of convolutional layers  $f_{cl}$ are copied to feature extractor $f_v$ and a CleanNet is trained to convergence. At step 3, the classifier are fine-tuned using the sample weights proposed by CleanNet. A similar alternating process can continue till the classifier stops improving. For more iterations of learning classifier, we fix the convolutional layers and only fine-tune the fully-connected layers.

\section{Experiments}
\subsection{Datasets}
\label{subsec:datasets}

\begin{table}
\begin{center}
\small
\begin{tabular}{|l|l|l|l|}
\hline
dataset & \#class & \#images & \#v-labels \\
\hline\hline
Food-101N & 101 & 310k/ - /25k & 55k/5k \\
Clothing1M & 14 & 1M/14k/10k & 25k/7k \\
WebVision & 1000 & 2.4M/50k/ - & 25k/ - \\
\hline
\end{tabular}
\end{center}
\caption{Datasets. \#images shows the numbers of images in train/val/test sets for classification (the train set is noisy labeled). \#v-labels shows the numbers of validation labels in train/val sets. }
\label{tb:datasets}
\end{table}

Table \ref{tb:datasets} lists the statistics of the datasets.

\noindent
\textbf{Food-101N:} We collect 310k images from Google, Bing, Yelp, and TripAdvisor using the Food-101 \cite{bossard2014food} taxonomy, and avoid foodspotting.com where the original Food-101 was collected. The estimated noisy class label accuracy is 80\%. We manually add 55k verification label for training and 5k for testing label noise detection. Image classification is evaluated on Food-101 test set.

\noindent
\textbf{Clothing1M \cite{xiao2015learning}:} Clothing1M is a public large-scale dataset designed for learning from noisy data with human supervision. It consists of 1M images with noisy class labels from 14 fashion classes. The estimated accuracy of class labels is 61.54\%. There are also three sets of images, with the size of 50k, 14k, 10k, respectively, which have correct class labels provided by human labelers -- we call them clean sets. There are some images overlap between the three clean sets and the  noisy set. For those overlapped images, we can then verify whether the noisy class label (as in the noisy set) is correct given the human labels on these images, and hence obtain verification labels for these images. Through this process, we obtain 25k and 7k verification labels for training and validation, respectively. The state of the art result of image classification on Clothing1M is reported in \cite{patrini2017making}.

\noindent
\textbf{WebVision\cite{li2017webvision}:} WebVision contains 2.4M noisy labeled images crawled from Flickr and Google using the ILSVRC taxonomy \cite{deng2009imagenet}. We conveniently verify noisy class labels using the Inception-ResNet-V2 model \cite{szegedy2017inception} pre-trained on ILSVRC. Noisy class label of an image is verified as relevant if it falls in top-5 predictions. Otherwise, the noisy class label is marked as mislabeled. We randomly obtain 250 ``pseudo-verification labels'' for each class for training. For evaluating image classification, we use 50k WebVision validation set and 50k ILSVRC 2012 validation set.

\subsection{Label Noise Detection}
\label{subsec:label-noise-detection}


We first evaluate CleanNet for the task of label noise detection. The label noise detection problem can be viewed as a binary classification problem for each class, and hence the results and comparisons are reported in average error rate over all the classes. We compare with the following categories of existing baseline methods:
\begin{itemize}
\setlength\itemsep{0em}
\item Supervised: Supervised methods learn a binary classification from verification labels for each class. We consider neural networks (2-layer MLP, used in \cite{yu2015lsun} for data construction), \textit{k}NN, SVM, label prop \cite{zhu2002learning}, and label spread \cite{zhou2004learning}. We also explored MLPs of more layers but 2-layer shows the best results.
\item Unsupervised: We consider DRAE \cite{xia2015learning}, the state of the art unsupervised outlier removal. Empirically, DRAE shows better results than one-class SVM \cite{scholkopf2001estimating}.
\item Weakly supervised: Like unsupervised method, weakly supervised methods do not require verification labels. We compare with a widely used classification filtering method: we train a CNN model on noisy data and predict top-K classes for each training image. An image is classified as relevant to its class label if the class is in top-K predictions. Otherwise, it is classified as mislabeled. K is selected on the validation set.
\end{itemize}

We provide two additional baselines: \textit{naive baseline} that treats all class labels as correct, and \textit{average baseline} that simply averages reference features as a class embedding vector and use query feature as a query embedding vector.

CleanNet and all the baselines depend on a CNN to extract image features. We fine-tune the ImageNet pre-trained ResNet-50 models \cite{he2016deep} on noisy data, same as step 1 in the alternating training scheme, and extract the \textit{pool5} layer as image features. Implementations of \textit{k}NN, SVM, label prop, and label spread are from scikit-learn \cite{pedregosa2011scikit}. We re-implemented DRAE and MLP in our experimentation.

In the following, we will evaluate CleanNet for label noise detection under two scenarios: \textbf{Full supervision}: verification labels in all classes are available for learning CleanNet; \textbf{Transfer learning}: only a fraction of classes contains verification labels for learning CleanNet.

\begin{table}[t]
\begin{center}
\small
\begin{tabular}{|l|c|c|}
\hline
& \multicolumn{2}{|c|}{average error rate} \\
\hline
method & Food-101N & Clothing1M \\
\hline\hline
naive baseline & 19.66 & 38.46\\
\hline
\multicolumn{3}{|c|}{supervised baselines} \\
\hline
MLP & 10.42  & 16.09\\
\textit{k}NN & 13.28  & 17.58 \\
SVM & 11.21  & 16.75\\
label prop \cite{zhu2002learning} & 13.24 & 17.81\\
label spread \cite{zhou2004learning} & 12.03  & 17.71\\
\hline
\multicolumn{3}{|c|}{weakly supervised baselines} \\
\hline
classification filtering & 16.60 & 23.55\\
\hline
\multicolumn{3}{|c|}{unsupervised baselines} \\
\hline
DRAE \cite{xia2015learning} & 18.70 & 38.95\\
average baseline & 16.20 & 30.56\\
\hline
\multicolumn{3}{|c|}{CleanNet (full supervision)} \\
\hline
CleanNet & \textbf{9.61} & \textbf{15.91}\\
CleanNet* & \textbf{6.99} & \textbf{15.77}\\
\hline
\end{tabular}
\end{center}
\caption{Label noise detection in terms of average error rate over all the classes (\%). CleanNet* denotes the results using image features extracted from the classifiers retrained with data cleansed by CleanNet.}
\label{tb:labelnoise}
\end{table}

\noindent
\textbf{Full supervision.} In Table \ref{tb:labelnoise}, we report the label noise detection results in terms of average error rate over all the classes. CleanNet gives error rate of 9.61\% on Food-101N and 15.91\% on Clothing1M. Comparing to MLP at 10.42\% on Food-101N and 16.09\% on Clothing1M, we validate that CleanNet performs similar to the best supervised baseline. Comparing to classification filtering at 16.60\% on Food-101N and 23.55\% on Clothing1M, the results demonstrate effectiveness of adding verification labels for human supervision for label noise detection. CleanNet* denotes the results of CleanNet using image features extracted from the classifiers retrained with data cleansed by CleanNet, and shows improvements (6.99\% on Food-101N and 15.77\% on Clothing1M). However, improvements become negligible with more iterations.

\begin{figure}[t]
\begin{center}
\includegraphics[width=1.0\linewidth]{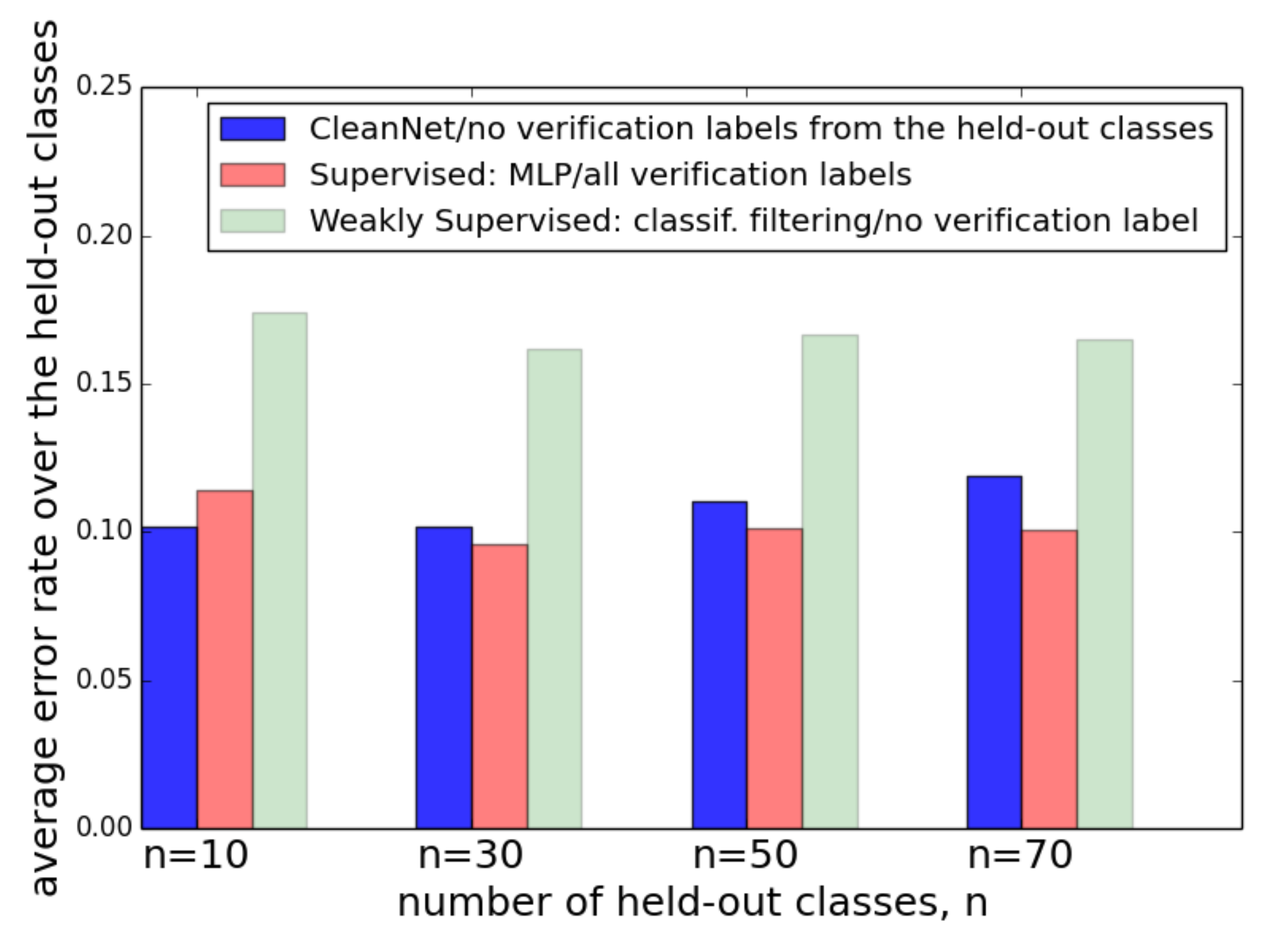}
\end{center}
\caption{Label noise detection on Food-101N with transfer learning. Verification labels in $n/101$ classes are held out for learning CleanNet, whereas MLP still uses all verification labels. Note that average error rate (\%) are ONLY evaluated on $n$ classes held out for CleanNet (so the numbers for MLP and classification filtering fluctuate for different $n$).}
\label{fig:leave-n-out-denoise}
\end{figure}

\noindent
\textbf{Transfer learning.} We choose Food-101N to demonstrate label noise detection with CleanNet under the setting of transfer learning, where verification labels in $n$ classes are held out for CleanNet (Lists of the held-out classes are available in the Food-101N dataset.). Here we also consider MLP that uses all verification labels and classification filtering that needs no verification labels. We ONLY evaluate the results on $n$ held-out classes to demonstrate the results on classes without explicit human supervision. The results are shown in Fig. \ref{fig:leave-n-out-denoise}. First, we observe that CleanNet can reduce label noise detection error rate on held-out classes where no human supervision available by 41.5\% relatively ($n=10$) compared to classification filtering. CleanNet consistently outperforms classification filtering, the weakly-supervised baseline. We also observe that the result of CleanNet with 50/101 classes held out (11.02\%) is still comparable to the result of MLP which is based on supervised learning (10.12\%).

\begin{table}
\begin{center}
\small
\begin{tabular}{|l|l|c|}
\hline
method & data & top-1 accuracy \\
\hline\hline
None & Food-101 & 81.67 \\
None & Food-101N & 81.44 \\
\hline
CleanNet, $w_{hard}$ & Food-101N  & 83.47 \\
CleanNet, $w_{soft}$ & Food-101N  & \textbf{83.95} \\
\hline
\end{tabular}
\end{center}
\caption{Image classification on Food-101N in terms of top-1 accuracy (\%). Verification labels in all classes are available. ``None'' denotes classifier without any method for label noise.}
\label{tb:f101n-classification}
\end{table}

\subsection{Learning Classifiers with Label Noise}
\label{subsec:image-recognition}
In this subsection, we present experiments for learning image classification models with label noise using the proposed CleanNet-based learning framework. Experimentation in this section is based on ResNet-50.

\begin{figure}[t]
\begin{center}
\includegraphics[width=1.0\linewidth]{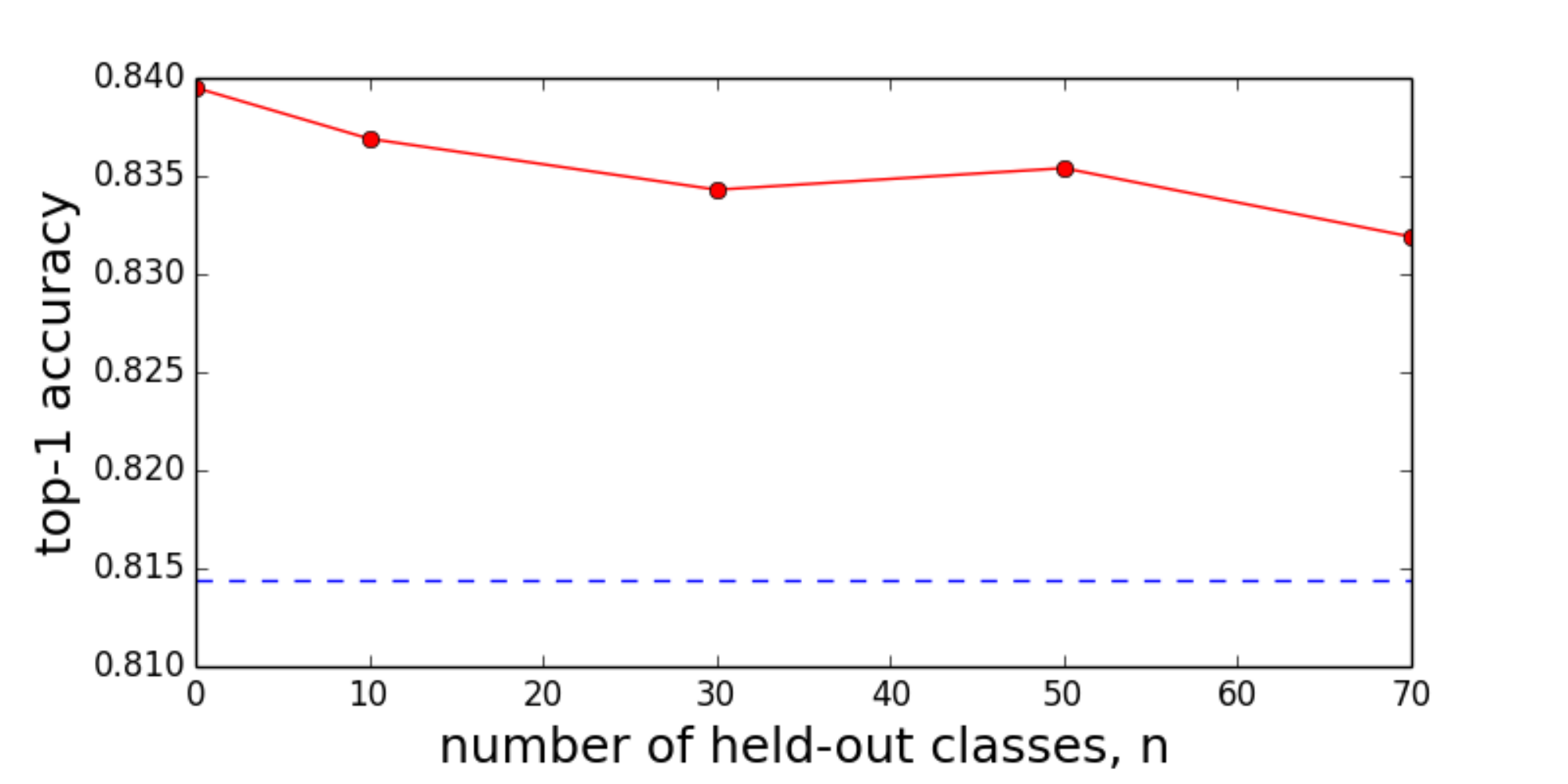}
\end{center}
\caption{Image classification on Food-101N in terms of top-1 accuracy (\%). Red line shows the results when verification labels in $n/101$ classes are held out for CleanNet. The blue dashed line shows the baseline without using CleanNet.}
\label{fig:f101n-da-classification}
\end{figure}

\begin{figure}[t]
\begin{center}
\includegraphics[width=1.0\linewidth]{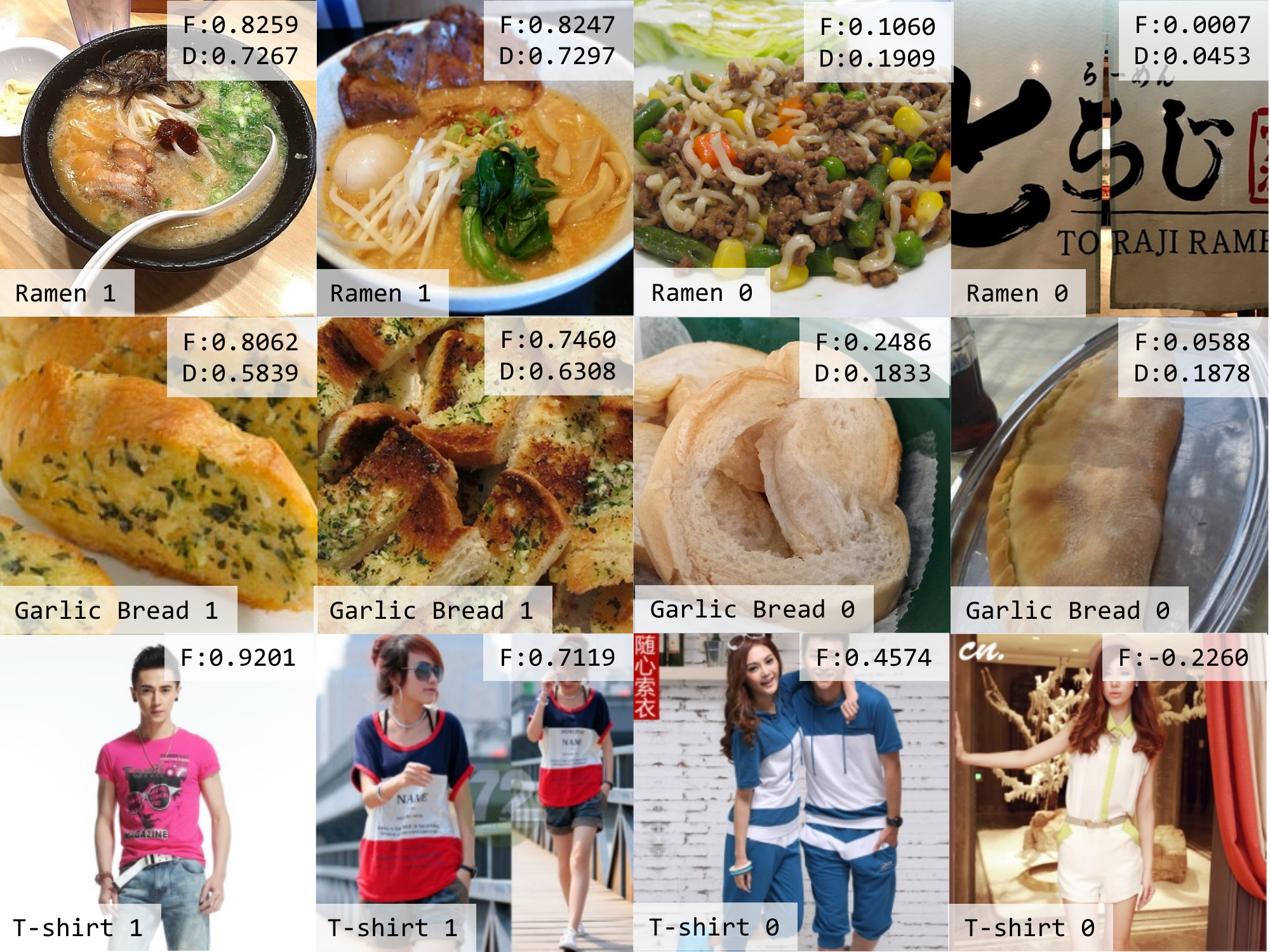}
\end{center}
\caption{Selected examples of CleanNet results on Food-101N and Clothing1M. ``F'' denotes cosine similarity predicted by model using verification labels in all classes. ``D'' denotes cosine similarity under transfer learning (50/101 classes are excluded for Food-101N, including ramen and garlic bread). Class names and verification labels are shown at bottom-left.}
\label{fig:denoise_vis}
\end{figure}

\noindent
\textbf{Experiments on Food-101N.} 
Table \ref{tb:f101n-classification} lists the results on Food-101N using verification labels in all classes. We observe that the performance of smooth soft weighting ($w_{soft}$) (83.95\%) without need for thresholding outperforms hard weighting ($w_{hard}$) (83.47\%). Fig. \ref{fig:f101n-da-classification} presents the results of image classification using the proposed CleanNet-based method when verification labels in $n$ classes are held out. For these $n$ held-out classes, the information needed for cleaning up the noisy class labels are transferred from other classes through CleanNet. It is observed that there are still 2.1\% and 1.75\% accuracy gain when 50/101 and 70/101 classes are held out. This validates that labeling effort on a small fraction of classes can still lead to significant gains.

Fig. \ref{fig:denoise_vis} shows examples of predictions by CleanNet. The cosine similarity score between the image and the reference set of its class is shown for each example. Because of transfer learning, CleanNet can assign reasonable scores to images from classes where no training images belonging to it are manually verified.

\begin{table}
\begin{center}
\small
\begin{tabular}{|l|l|l|l|l|}
\hline
\# & method & data & pretrained & top-1 \\
\hline\hline
1 & None \cite{patrini2017making} & 1M noisy & ImageNet & 68.94 \\
2 & None \cite{patrini2017making} & 50k clean & ImageNet & 75.19 \\
\hline
3 & loss correct. \cite{patrini2017making} & 1M noisy & ImageNet & 69.84 \\
4 & None \cite{patrini2017making} & 50k clean & \#3 model & \textbf{80.38}$^\dagger$ \\
\hline
5 & CleanNet,$w_{hard}$ & 1M noisy & ImageNet & 74.15 \\
6 & CleanNet,$w_{soft}$ & 1M noisy & ImageNet & \textbf{74.69} \\
7 & None & 50k clean & \#6 model & \textbf{79.90} \\
\hline
\end{tabular}
\end{center}
\caption{Image classification on Clothing1M in terms of top-1 accuracy (top-1)(\%). ``None'' denotes classifier without any method for label noise.  $^\dagger$: the result is not directly comparable to ours (See Sec. \ref{subsec:image-recognition} for more details).}
\label{tb:c1m-classification}
\end{table}

\noindent
\textbf{Experiments on Clothing1M.} For Clothing1M, we consider the state of the art result reported in \cite{patrini2017making}, which also used ResNet-50. \cite{patrini2017making} used the part of data in Clothing1M that has both noisy and correct class labels to estimate confusion among classes and modeled this information in loss function. Since we only compare the noisy class label to the correct class label for an image to verify whether the noisy class label is correct, we lose the label confusion information, and thus these numbers are not directly comparable. However, labeling the correct classes like Clothing1M (only 14 classes) is not scalable in number of classes because having labeling workers select from a large number of classes is time-consuming and unlikely to be accurate.

Table \ref{tb:c1m-classification} lists the results of image classification using verification labels in all classes. Using CleanNet significantly improves the accuracy from 68.94\% (\#1) to 74.69\% (\#6) on 1M noisy training data. We also follow \cite{patrini2017making} to fine-tune the best model trained on 1M noisy set on the 50k clean training set. Our proposed method achieves 79.90\%, which is comparable to the state of the art 80.38\% reported in \cite{patrini2017making} which benefits from the extra label confusion information.

\begin{table}
\begin{center}
\small
\begin{tabular}{|l|l|}
\hline
verification & definition \\
\hline\hline
every-image & verification labels for every image\\
\hline
all-1000 & all 1000 classes\\
\hline
semantic-308 & \parbox[t]{5.6cm}{308 classes selected from each group of classes that share a common second-level hypernym in WordNet \cite{miller1995wordnet}}\\
\hline
random-308 & random selected 308 classes\\
\hline
random-118 & random selected 118 classes\\
\hline
dogs-118 & 118 dog classes\\
\hline
\end{tabular}
\end{center}
\caption{Verification conditions: selecting different classes for adding verification labels. Other than every-image, all other conditions have only 250 verification labels in each class.}
\label{tb:webvision_veri}
\end{table}

\begin{table}
\begin{center}
\small
\begin{tabular}{|l|l|c|c|}
\hline
& &\multicolumn{2}{|c|}{val acc top-1(top-5)} \\
\hline
method & verification & WebVision & ILSVRC \\
\hline\hline
baseline & - & 67.76(85.75) & 58.88(79.76) \\
upper bnd & every-image & 70.31(87.77) & 63.42(84.59) \\
\hline
CleanNet & all-1000& 69.14(86.73)  & 61.03(82.01)\\
CleanNet & semantic-308 & 68.96(86.64) & 60.48(81.40)\\
CleanNet & random-308 & 68.89(86.61)  & 60.27(81.27)\\
CleanNet & random-118 & 68.50(86.51)  & 60.16(81.05)\\
CleanNet & dogs-118  & 68.33(86.04)  & 59.43(80.22)\\
\hline
\end{tabular}
\end{center}
\caption{Image classification on WebVision in terms of top-1 and top-5 accuracy (\%). The models are trained WebVision training set and tested on WebVision and ILSVRC validation sets under various verification conditions.}
\label{tb:webvision}
\end{table}

\noindent
\textbf{Experiments on WebVision.} As opposed to Food-101N and Clothing1M which are fine-grained tasks, WebVision experiments sheds light on general image classification at very large scale. As mentioned in Sec. \ref{subsec:datasets}, the pseudo-verification labels are model-based so that we can obtain for all images. This property allows us to explore how to select classes for adding verification labels and compare to the upper bound scenario where all noisy class labels are verified without any cost. We define how to add verification labels as ``verification conditions'', listed in Table \ref{tb:webvision_veri}. Table \ref{tb:webvision} shows the experimental results using CleanNet and soft weighting ($w_{soft}$). We observe that verifying every image (every-image) improves the top-1 accuracy from 67.76\% to 70.31\% on the WebVision validation set. With only 3.20\% and 1.2\% images verified, semantic-308 and random-118 give 47\% and 29\% of the performance gain of every-image on the WebVision validation set respectively. Note that we only include 250 verification labels for each class for all experiments using CleanNet. The results again confirm that labeling on a fraction of classes is effective because of transfer learning by CleanNet.



\section{Conclusion}
In this work, we highlighted the difficulties of having both scalability and effectiveness of human supervision for label noise detection and classification learning from noisy data. We introduced CleanNet as a transfer learning approach to reconcile the issue by transferring supervised information of transferring the correctness of labels to classes without explicit human supervision. We empirically evaluate our proposed methods on both general and fine-grained image classification datasets. The results show that CleanNet outperforms methods using no human supervision by a large margin when small fraction of classes is manually verified. It also matches existing methods that require extensive human supervision when sufficient classes are manually verified. We believe this work creates a novel paradigm that efficiently utilizes human supervision to better address label noise in large-scale image classification tasks.

\section*{Acknowledgement} The authors thank Xi Chen, Yu-Hsiang Bosco Chiu, Yandong Guo and Po-Sen Huang for their thoughtful feedbacks and discussions. Thanks also to Li Huang and Arun Sacheti for helping develop the Food-101N dataset. 

{\small
\bibliographystyle{ieee}
\bibliography{egbib}

\begin{thebibliography}{10}\itemsep=-1pt

\bibitem{azadi2016auxiliary}
S.~Azadi, J.~Feng, S.~Jegelka, and T.~Darrell.
\newblock Auxiliary image regularization for deep {CNN}s with noisy labels.
\newblock In {\em ICLR}, 2016.

\bibitem{bossard2014food}
L.~Bossard, M.~Guillaumin, and L.~Van~Gool.
\newblock Food-101--mining discriminative components with random forests.
\newblock In {\em ECCV}, 2014.

\bibitem{chen2015webly}
X.~Chen and A.~Gupta.
\newblock Webly supervised learning of convolutional networks.
\newblock In {\em ICCV}, 2015.

\bibitem{dean2013fast}
T.~Dean, M.~A. Ruzon, M.~Segal, J.~Shlens, S.~Vijayanarasimhan, and J.~Yagnik.
\newblock Fast, accurate detection of 100,000 object classes on a single
  machine.
\newblock In {\em CVPR}, 2013.

\bibitem{deng2009imagenet}
J.~Deng, W.~Dong, R.~Socher, L.-J. Li, K.~Li, and L.~Fei-Fei.
\newblock {ImageNet}: A large-scale hierarchical image database.
\newblock In {\em CVPR}, 2009.

\bibitem{fergus2010learning}
R.~Fergus, L.~Fei-Fei, P.~Perona, and A.~Zisserman.
\newblock Learning object categories from internet image searches.
\newblock {\em Proceedings of the IEEE}, 98(8):1453--1466, 2010.

\bibitem{frenay2014classification}
B.~Fr{\'e}nay and M.~Verleysen.
\newblock Classification in the presence of label noise: a survey.
\newblock {\em IEEE Transactions on Neural Networks and Learning Systems},
  25(5):845--869, 2014.

\bibitem{frome2013devise}
A.~Frome, G.~S. Corrado, J.~Shlens, S.~Bengio, J.~Dean, T.~Mikolov, et~al.
\newblock Devise: A deep visual-semantic embedding model.
\newblock In {\em NIPS}, 2013.

\bibitem{he2016deep}
K.~He, X.~Zhang, S.~Ren, and J.~Sun.
\newblock Deep residual learning for image recognition.
\newblock In {\em CVPR}, 2016.

\bibitem{hinton2006reducing}
G.~E. Hinton and R.~R. Salakhutdinov.
\newblock Reducing the dimensionality of data with neural networks.
\newblock {\em Science}, 313(5786):504--507, 2006.

\bibitem{krause2016unreasonable}
J.~Krause, B.~Sapp, A.~Howard, H.~Zhou, A.~Toshev, T.~Duerig, J.~Philbin, and
  L.~Fei-Fei.
\newblock The unreasonable effectiveness of noisy data for fine-grained
  recognition.
\newblock In {\em ECCV}, 2016.

\bibitem{krizhevsky2012imagenet}
A.~Krizhevsky, I.~Sutskever, and G.~E. Hinton.
\newblock {ImageNet} classification with deep convolutional neural networks.
\newblock In {\em NIPS}, 2012.

\bibitem{li2017webvision}
W.~Li, L.~Wang, W.~Li, E.~Agustsson, and L.~Van~Gool.
\newblock Webvision database: Visual learning and understanding from web data.
\newblock {\em arXiv preprint arXiv:1708.02862}, 2017.

\bibitem{li2017learning}
Y.~Li, J.~Yang, Y.~Song, L.~Cao, J.~Luo, and J.~Li.
\newblock Learning from noisy labels with distillation.
\newblock In {\em ICCV}, 2017.

\bibitem{lin2014microsoft}
T.-Y. Lin, M.~Maire, S.~Belongie, J.~Hays, P.~Perona, D.~Ramanan,
  P.~Doll{\'a}r, and C.~L. Zitnick.
\newblock Microsoft {COCO}: Common objects in context.
\newblock In {\em ECCV}, 2014.

\bibitem{liu2016coupled}
M.-Y. Liu and O.~Tuzel.
\newblock Coupled generative adversarial networks.
\newblock In {\em NIPS}, 2016.

\bibitem{liu2014unsupervised}
W.~Liu, G.~Hua, and J.~R. Smith.
\newblock Unsupervised one-class learning for automatic outlier removal.
\newblock In {\em CVPR}, 2014.

\bibitem{miller1995wordnet}
G.~A. Miller.
\newblock {WordNet}: a lexical database for {English}.
\newblock {\em Communications of the ACM}, 38(11):39--41, 1995.

\bibitem{nettleton2010study}
D.~F. Nettleton, A.~Orriols-Puig, and A.~Fornells.
\newblock A study of the effect of different types of noise on the precision of
  supervised learning techniques.
\newblock {\em Artificial Intelligence Review}, 33(4):275--306, 2010.

\bibitem{patrini2017making}
G.~Patrini, A.~Rozza, A.~Krishna~Menon, R.~Nock, and L.~Qu.
\newblock Making deep neural networks robust to label noise: A loss correction
  approach.
\newblock In {\em CVPR}, 2017.

\bibitem{pedregosa2011scikit}
F.~Pedregosa, G.~Varoquaux, A.~Gramfort, V.~Michel, B.~Thirion, O.~Grisel,
  M.~Blondel, P.~Prettenhofer, R.~Weiss, V.~Dubourg, et~al.
\newblock Scikit-learn: Machine learning in python.
\newblock {\em Journal of Machine Learning Research}, 12(Oct):2825--2830, 2011.

\bibitem{rolnick2017deep}
D.~Rolnick, A.~Veit, S.~Belongie, and N.~Shavit.
\newblock Deep learning is robust to massive label noise.
\newblock {\em arXiv preprint arXiv:1705.10694}, 2017.

\bibitem{salvador2017learning}
A.~Salvador, N.~Hynes, Y.~Aytar, J.~Marin, F.~Ofli, I.~Weber, and A.~Torralba.
\newblock Learning cross-modal embeddings for cooking recipes and food images.
\newblock In {\em CVPR}, 2017.

\bibitem{scholkopf2001estimating}
B.~Sch{\"o}lkopf, J.~C. Platt, J.~Shawe-Taylor, A.~J. Smola, and R.~C.
  Williamson.
\newblock Estimating the support of a high-dimensional distribution.
\newblock {\em Neural Computation}, 13(7):1443--1471, 2001.

\bibitem{schroff2011harvesting}
F.~Schroff, A.~Criminisi, and A.~Zisserman.
\newblock Harvesting image databases from the web.
\newblock {\em IEEE Transactions on Pattern Analysis and Machine Intelligence},
  33(4):754--766, 2011.

\bibitem{socher2013zero}
R.~Socher, M.~Ganjoo, C.~D. Manning, and A.~Ng.
\newblock Zero-shot learning through cross-modal transfer.
\newblock In {\em NIPS}, 2013.

\bibitem{sukhbaatar2014training}
S.~Sukhbaatar, J.~Bruna, M.~Paluri, L.~Bourdev, and R.~Fergus.
\newblock Training convolutional networks with noisy labels.
\newblock {\em arXiv preprint arXiv:1406.2080}, 2014.

\bibitem{szegedy2017inception}
C.~Szegedy, S.~Ioffe, V.~Vanhoucke, and A.~A. Alemi.
\newblock Inception-v4, inception-resnet and the impact of residual connections
  on learning.
\newblock In {\em AAAI}, 2017.

\bibitem{thongkam2008support}
J.~Thongkam, G.~Xu, Y.~Zhang, and F.~Huang.
\newblock Support vector machine for outlier detection in breast cancer
  survivability prediction.
\newblock In {\em Asia-Pacific Web Conference}, pages 99--109. Springer, 2008.

\bibitem{tsai2017learning}
Y.-H.~H. Tsai, L.-K. Huang, and R.~Salakhutdinov.
\newblock Learning robust visual-semantic embeddings.
\newblock In {\em ICCV}, 2017.

\bibitem{tzeng2017adversarial}
E.~Tzeng, J.~Hoffman, K.~Saenko, and T.~Darrell.
\newblock Adversarial discriminative domain adaptation.
\newblock In {\em CVPR}, 2017.

\bibitem{veit2017learning}
A.~Veit, N.~Alldrin, G.~Chechik, I.~Krasin, A.~Gupta, and S.~Belongie.
\newblock Learning from noisy large-scale datasets with minimal supervision.
\newblock In {\em CVPR}, 2017.

\bibitem{vinyals2016matching}
O.~Vinyals, C.~Blundell, T.~Lillicrap, k.~kavukcuoglu, and D.~Wierstra.
\newblock Matching networks for one shot learning.
\newblock In {\em NIPS}, 2016.

\bibitem{xia2015learning}
Y.~Xia, X.~Cao, F.~Wen, G.~Hua, and J.~Sun.
\newblock Learning discriminative reconstructions for unsupervised outlier
  removal.
\newblock In {\em ICCV}, 2015.

\bibitem{xiao2015learning}
T.~Xiao, T.~Xia, Y.~Yang, C.~Huang, and X.~Wang.
\newblock Learning from massive noisy labeled data for image classification.
\newblock In {\em CVPR}, 2015.

\bibitem{yang2016hierarchical}
Z.~Yang, D.~Yang, C.~Dyer, X.~He, A.~Smola, and E.~Hovy.
\newblock Hierarchical attention networks for document classification.
\newblock In {\em NAACL HLT}, 2016.

\bibitem{yu2015lsun}
F.~Yu, A.~Seff, Y.~Zhang, S.~Song, T.~Funkhouser, and J.~Xiao.
\newblock {LSUN}: Construction of a large-scale image dataset using deep
  learning with humans in the loop.
\newblock {\em arXiv preprint arXiv:1506.03365}, 2015.

\bibitem{zhou2017places}
B.~Zhou, A.~Lapedriza, A.~Khosla, A.~Oliva, and A.~Torralba.
\newblock Places: A 10 million image database for scene recognition.
\newblock {\em IEEE Transactions on Pattern Analysis and Machine Intelligence},
  2017.

\bibitem{zhou2004learning}
D.~Zhou, O.~Bousquet, T.~N. Lal, J.~Weston, and B.~Sch{\"o}lkopf.
\newblock Learning with local and global consistency.
\newblock In {\em NIPS}, 2004.

\bibitem{zhu2002learning}
X.~Zhu and Z.~Ghahramani.
\newblock Learning from labeled and unlabeled data with label propagation.
\newblock {\em Technical Report CMU-CALD-02-107}, 2002.

\bibitem{zhuang2017attend}
B.~Zhuang, L.~Liu, Y.~Li, C.~Shen, and I.~Reid.
\newblock Attend in groups: a weakly-supervised deep learning framework for
  learning from web data.
\newblock In {\em CVPR}, 2017.

\end{thebibliography}
}

\end{document}